\documentclass[a4paper,fleqn]{cas-dc}

\usepackage[numbers,longnamesfirst]{natbib}
\usepackage{graphicx}
\usepackage{subcaption} 
\usepackage{amsmath}
\usepackage{amssymb}
\usepackage{multirow}
\usepackage{tabularx}
\usepackage{makecell}
\usepackage{booktabs}
\usepackage{array}

\def\tsc#1{\csdef{#1}{\textsc{\lowercase{#1}}\xspace}}
\tsc{WGM}
\tsc{QE}
\tsc{EP}
\tsc{PMS}
\tsc{BEC}
\tsc{DE}

\begin{document}

\let\WriteBookmarks\relax
\def\floatpagepagefraction{1}
\def\textpagefraction{.001}
\let\printorcid\relax

\shorttitle{SEMPose: A Single End-to-end Network for Multi-object Pose Estimation}

\shortauthors{Xin Liu et~al.}

\title [mode = title]{SEMPose: A Single End-to-end Network for Multi-object Pose Estimation}                      

\tnotetext[1]{This work was supported in part by the National Natural Science Foundation of China under Grant 62273226 and Grant 61873162.}


\author[1]{Xin Liu}
\ead{liu.xin@sjtu.edu.cn}

\fnmark[1]

\credit{Conceptualization, Methodology, Coding, Writing original draft}

\affiliation[1]{organization={Department of Automation, Shanghai Jiao Tong University},
    city={Shanghai},
    postcode={200240}, 
    country={China}}

\author[1]{Hao Wang}
\credit{Coding, Writing, Reviewing, Editing}
\ead{wh631q@sjtu.edu.cn}

\author[1]{Shibei Xue}
\credit{Supervision, Writing, Reviewing, Editing}
\cormark[1]
\ead{shbxue@sjtu.edu.cn}

\author[2]{Dezong Zhao}
\credit{Supervision, Writing, Reviewing, Editing}
\ead{Dezong.Zhao@glasgow.ac.uk}

\affiliation[2]{organization={James Watt School of Engineering, University of Glasgow},
    city={Glasgow},
    postcode={G12 8QQ}, 
    country={United Kingdom}}

\cortext[cor1]{Corresponding author: Shibei Xue}

\begin{abstract}
In computer vision, estimating the six-degree-of-freedom pose from an RGB image is a fundamental task. However, this task becomes highly challenging in multi-object scenes. Currently, the best methods typically employ an indirect strategy, which identifies 2D and 3D correspondences, and then solves with the Perspective-n-Points method. Yet, this approach cannot be trained end-to-end. Direct methods, on the other hand, suffer from lower accuracy due to challenges such as varying object sizes and occlusions. To address these issues, we propose SEMPose, an end-to-end multi-object pose estimation network. SEMPose utilizes a well-designed texture-shape guided feature pyramid network, effectively tackling the challenge of object size variations. Additionally, it employs an iterative refinement head structure, progressively regressing rotation and translation separately to enhance estimation accuracy. During training, we alleviate the impact of occlusion by selecting positive samples from visible parts. Experimental results demonstrate that SEMPose can perform inference at 32 FPS without requiring inputs other than the RGB image. It can accurately estimate the poses of multiple objects in real time, with inference time unaffected by the number of target objects. On the LM-O and YCB-V datasets, our method outperforms other RGB-based single-model methods, achieving higher accuracy. Even when compared with multi-model methods and approaches that use additional refinement, our results remain competitive.
\end{abstract}

\begin{keywords} 
Multi-object pose estimation \sep
End-to-end network  \sep
Feature fusion \sep
Positive sample selection   \sep
\end{keywords}

\maketitle

\section{Introduction}
In the field of machine vision, six degrees of freedom (6D) pose estimation plays a crucial role. This technology can measure an object's position (coordinates along the x, y, and z axes) and orientation (roll, pitch, and yaw angles). 
Therefore, it can help robots accurately understand objects' spatial posture, which is crucial in robots grasping, moving, or manipulating objects\cite{he2022deep}. During these processes, it is common to encounter scenes with multiple objects to estimate. These objects often vary in size and may occlude each other. However,  despite significant advancements in 6D pose estimation technology, existing methods still struggle to effectively handle the multi-object scenes. 

Specifically, in recent years, with the deepening research in deep learning \cite{he2016deep,lin2017fpn,tang2023rethinking}, methods for 6D pose estimation have been continuously emerging\cite{li2019cdpn,labbe2020cosypose,wang2021gdr,xiangposecnn,peng2019pvnet,jantos2023poet}. At the same time, physically-based rendering techniques have narrowed the gap between synthetic and real images, improving the model's generalization ability in real-world scenarios \cite{denninger2020blenderproc}. In this context, deep learning-based methods have made significant advancements in the task of 6D pose estimation, surpassing traditional methods based on point-to-point features in terms of both accuracy and speed \cite{sundermeyer2023bop}. Among these methods, the most direct approach is to directly regress the 6D pose of objects from input images, without the need for additional steps or models for feature point detection or matching \cite{wang2021gdr,xiangposecnn,jantos2023poet,bukschat2020efficientpose}. These methods are suitable for both single-object and multi-object scenarios. And they are easy to deploy and train. However, they still lag behind state-of-the-art methods in terms of accuracy due to occlusions and varying object sizes\cite{thalhammer2023open,hai2023rigidity}. Indirect methods achieve higher accuracy by utilizing the idea of correspondence \cite{guan2023hrpose,guan2021high,zakharov2019dpod,hu2019segmentation,hodan2020epos}. These methods first predict the key points of objects in 2D images and then match them with corresponding 3D key points. Finally, these correspondences are input into the Perspective-n-Points (PnP) algorithm or RANSAC algorithm to obtain the object's pose. This two-stage approach has improved accuracy. However, these methods also have drawbacks: on the one hand, due to the non-differentiable nature of PnP and RANSAC, these methods cannot be directly trained and deployed end-to-end; on the other hand, even if only one object is processed, the iterative process of RANSAC can be very time-consuming\cite{hodan2020epos}.

In addition, there are some other challenges in existing methods when it comes to multi-object pose estimation tasks. On the one hand, existing methods mostly design networks for specific types of objects, making it difficult to generalize to other types of objects. Therefore, to improve accuracy on the YCB-V dataset, many methods train 21 pose estimation models for 21 different types of objects \cite{peng2019pvnet,xu2024rnnpose}. This means high system complexity and resource consumption when deployed in practice. On the other hand, in multi-object scenarios, there are differences in size and scale between objects, which can lead to imbalance and bias during network training\cite{thalhammer2023open}.

To address these issues, we propose SEMPose, a \textbf{S}ingle \textbf{E}nd-to-end network for \textbf{M}ulti-object \textbf{P}ose
estimation. Our network requires no additional information besides RGB images, such as 3D models\cite{lin2024sam}, depth images\cite{lin2024hipose}, object symmetry information\cite{billings2019silhonet}, or ground-truth RoI information\cite{jantos2023poet}.

Specifically, we adopt a backbone-neck-head structure that enables the entire network to be trained end-to-end. To address the issue of varying object scales in multi-object scenes, we designed a texture-shape guided feature pyramid structure to hierarchically capture the fused features of objects with different sizes. For the issue of object occlusions, we use the unoccluded parts of objects to guide positive sample selection during sampling. What's more, to improve the accuracy of pose prediction, we employ different strategies for predicting rotation and translation, and design rotation iteration head and translation iteration head. This enables our network to learn coordinate features closely related to translation and fully utilize contextual information for a more comprehensive perspective in prediction. 

Overall, our contributions are as follows:
\begin{itemize}
	\item We propose an end-to-end pose estimation network that can simultaneously handle multiple objects. Our approach requires only one model for multiple objects, and the runtime is independent of the number of objects. Additionally, we train the network using only raw RGB data.
	\item We propose a texture-shape guided feature pyramid structure that can handle objects of varying sizes and enhance feature extraction. Additionally, we decouple the prediction of rotation and translation, and propose iteration heads. This enables our network to improve the accuracy of rotation and translation predictions. Compared to GDR-Net, we reduce the average translation estimation error by 2.47cm and the average rotation estimation error by $8.87^{\circ}$.
        \item In methods using only RGB images, we achieve state-of-the-art performance on the LM-O and YCB-V datasets. Even compared to multi-model methods and those utilizing 3D models, SEMPose can achieve similar or even superior performance.
\end{itemize}

\section{Related Works}
\subsection{Indirect Methods.}
The mainstream methods for 6D pose estimation adopt an indirect regression approach. Some of these methods \cite{peng2019pvnet,guan2023hrpose,guan2021high,zakharov2019dpod,hu2019segmentation,hodan2020epos} are based on correspondence principles. They first detect feature points of the target object in an image, then match these feature points with corresponding features in a known model. Subsequently, they utilize algorithms such as RANSAC or Perspective-n-points (PnP) to compute the 6D pose of the target object based on the matched point pairs. 
For example, PVNet \cite{peng2019pvnet} predicts pixel-level vectors pointing to key points, then uses these vectors to determine the key points' positions through a RANSAC-based voting mechanism. Finally, PnP is used for calculation.
Some methods \cite{guan2023hrpose,guan2021high} predict belief maps and affinity maps for each object category. Belief maps represent the likelihood of keypoint locations, while affinity maps represent the correlation between 3D bounding box vertices and the corresponding center point. Then, 2D key points are extracted from the predicted belief maps, and finally, the PnP algorithm is used to calculate the pose.
 Although these methods can effectively reduce the impact of occlusions, the non-differentiable nature of RANSAC/PnP prevents end-to-end training and deployment; moreover, an increase in the number of objects in the scene significantly raises the demand for computational resources.

 Another group of methods is based on the idea of template matching\cite{lin2024sam,labbe2023megapose}. These methods first construct a 2D template database using a 3D model, and then compare the input image with all templates in the database using a sliding window algorithm to estimate the pose. The accuracy of this approach improves with an increase in the number of templates, but it also has several significant drawbacks: firstly, obtaining accurate 3D models can be challenging in practical applications; secondly, as the number of objects to be estimated in the scene increases, the size of the template database will also increase, leading to a significant decrease in processing speed; thirdly, changes in texture due to occlusions or lighting can greatly decrease accuracy.
 Overall, these indirect methods exhibit several limitations in scenarios involving multiple objects.
 \begin{figure*}[t]
  \centering
  \includegraphics[width=\linewidth]{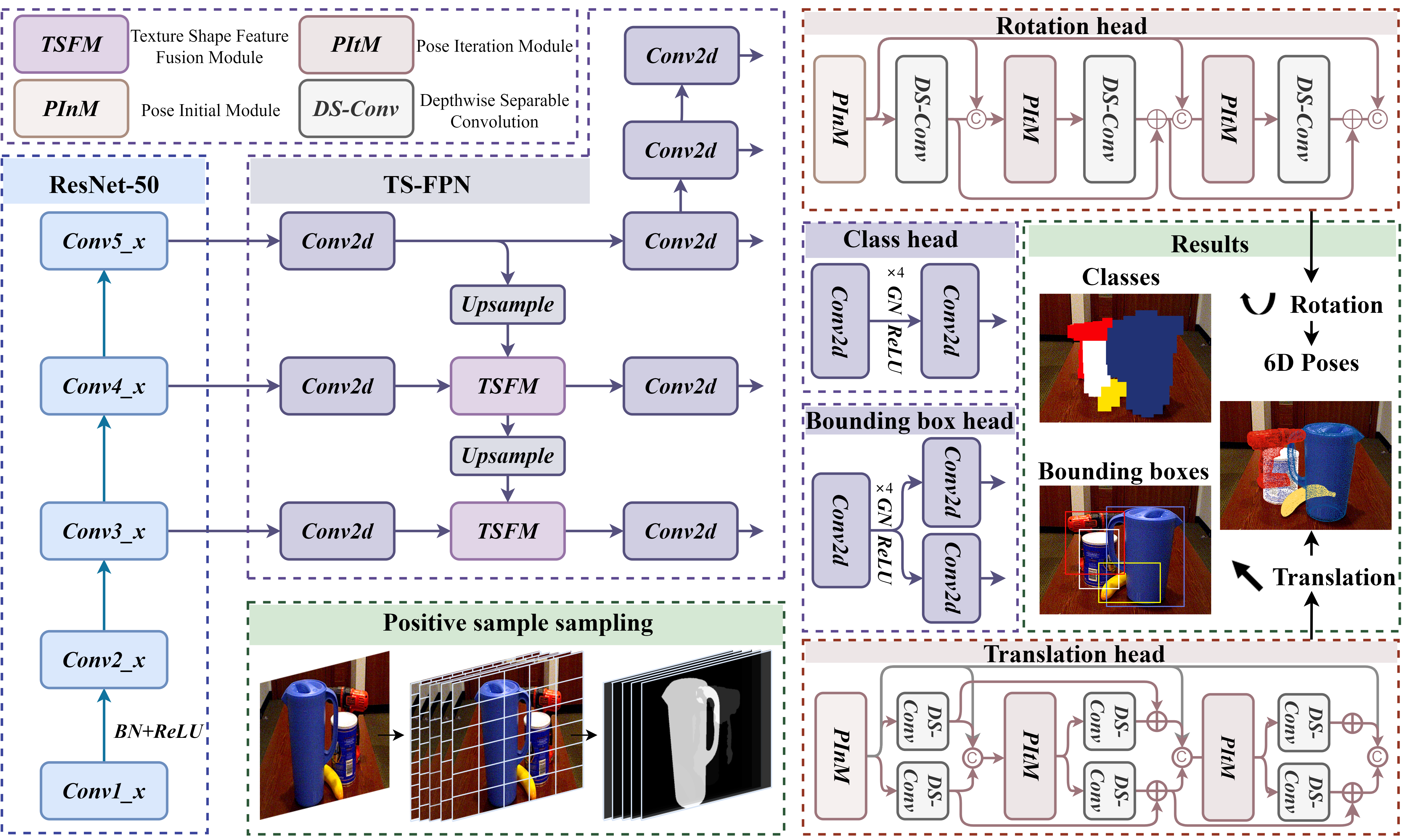}
  \caption{\textbf{Framework of SEMPose.} Given an input RGB image \textit{I}, SEMPose first uses Resnet50 to extract basic features. Then, SEMPose employs a Texture-Shape Guided Feature Pyramid Network (TS-FPN) to fuse these features and produce feature maps at five different scales. Finally, SEMPose uses four heads to predict the categories, bounding boxes, rotations, and translations of the objects. After post-processing, the poses are obtained. Additionally, the network's training process relies on a positive sample sampling strategy guided by the visible parts.}
  \label{overview}
\end{figure*}

\subsection{Direct Methods.}
The most straightforward way to estimate 6D pose is to directly regress the 6D pose\cite{labbe2020cosypose,wang2021gdr,xiangposecnn,jantos2023poet,bukschat2020efficientpose}. For instance, some methods \cite{labbe2020cosypose,xiangposecnn,bukschat2020efficientpose} evaluate the matching loss by calculating the average squared distance between points on the model under the ground truth pose and the estimated pose. This approach assesses the discrepancies in model points across the two poses to enhance the accuracy of 6D pose estimation.
PoET\cite{jantos2023poet} utilizes a neural network to generate multi-scale feature maps from monocular RGB images and detect objects. These features and bounding boxes are then processed through a Transformer, ultimately predicting the 6D pose of each object using separate rotation and translation heads.
GDR-Net\cite{wang2021gdr} first predicts intermediate geometric feature maps, including dense correspondence maps and surface area attention maps. A 2D convolutional Patch-PnP module then directly regresses the 6D pose from these geometric features.

 These methods are mostly end-to-end, which is beneficial for training and deployment. However, because convolutional neural networks directly regress the 6D pose from images, they face challenges in learning varying degrees of translation dependencies and struggle with effectively handling occlusions\cite{liu2018coordconv,hai2023rigidity}. Therefore, there is significant room for improvement in the accuracy of these methods.

\section{Proposed Method}
Given an RGB image \textit{I} containing \textit{N} objects $\mathcal{O} =\left\{\mathcal{O} _i\mid i=1,2,\cdots ,N\right\}$, our goal is to simultaneously estimate their 6D poses $
\mathbf{P}_{i\,\,}=\left[ \mathbf{R}_i|\mathbf{t}_i \right] , i=1,2,\cdots ,N$, where, $\mathbf{R}_i\in SO\left( 3 \right) $ is the rotation matrix, and $\mathbf{t}_i\in \mathbb{R} ^3$ is the translation vector. The entire 6D pose $\mathbf{P}_{i\,\,}$ represents a rigid transformation from the camera coordinate system to the object coordinate system.

Figure \ref{overview} provides an overview schematic of our proposed method. Our SEMPose consists of three parts: the backbone, neck, and heads. The core components include our designed TS-FPN, pose estimation heads, and a positive sample selection strategy guided by visible parts.

In the following, we will first (Sec. \ref{sec:tsfpn}) provide a detailed introduction to our TS-FPN. Then (Sec. \ref{sec:6d}), we will explain our regression strategy for rotation and translation. Following this (Sec. \ref{sec:heads}), we will describe how we designed the network to implement this strategy. Finally (Sec. \ref{sec:sample}), we will introduce our network's approach to handling occlusion, specifically by sampling from the visible parts.

\subsection{Texture-Shape Guided Feature Pyramid Network}
\label{sec:tsfpn}
In neural networks, lower layers typically capture basic features of input images, such as edges, colors, and texture information. These features vary with the rotation of objects. Therefore, by capturing these high-frequency details, neural networks can better understand the orientation of objects, leading to more accurate estimates of rotation. As the network deepens, the feature maps abstract the input data more deeply, depicting more complex edge structures and specific shapes. This low-frequency information often reflects the overall shape and larger structural features of objects, which is crucial for estimating the translation. This is because the global shape and position information of objects help the network determine the center and position of the object. For example, when an object moves, changes are mainly related to low-frequency information, while high-frequency information such as edges and textures remains relatively stable.

\begin{figure}
  \centering
  \includegraphics[width=\linewidth]{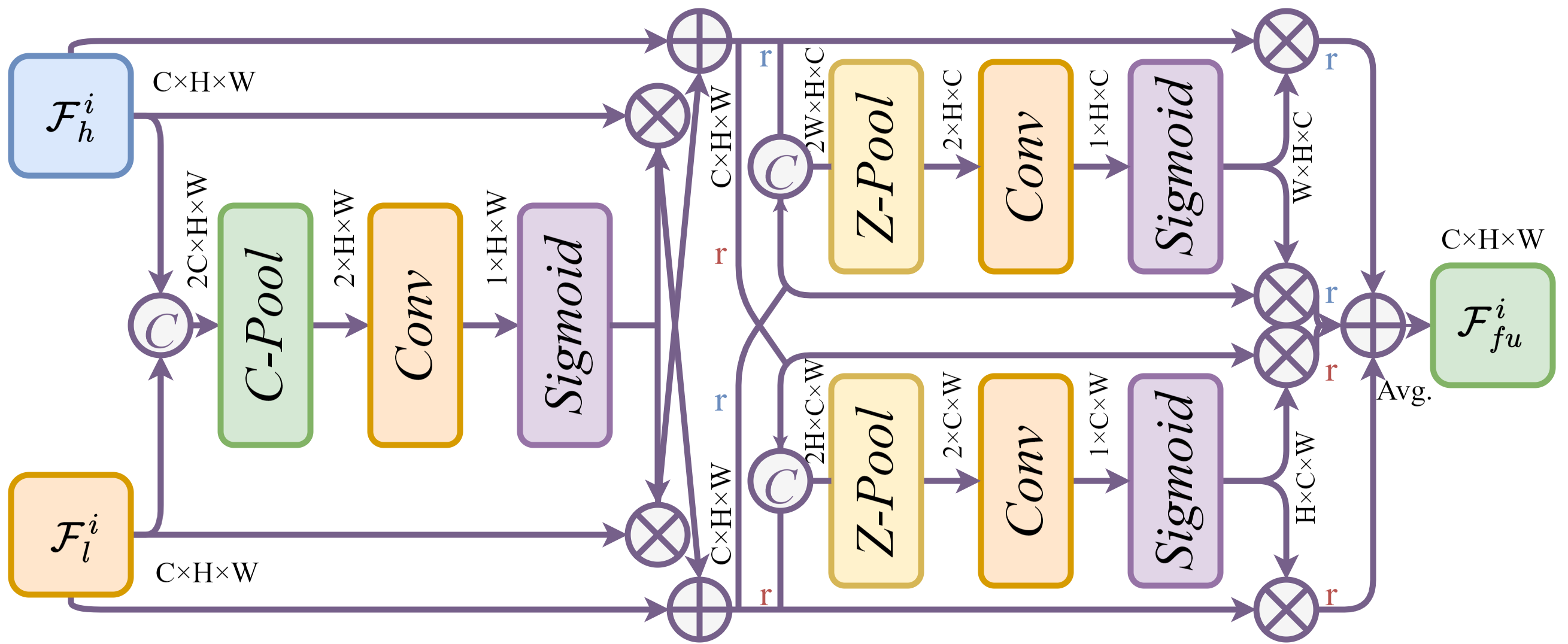}
  \caption{The structure diagram of texture shape feature fusion module.}
  \label{fig:TSFM}
\end{figure}

Traditional Feature Pyramid Networks\cite{lin2017fpn}, by integrating the high-resolution information from lower levels with the semantically rich information from higher levels, can better handle targets of different scales. This has achieved good results in the field of object detection, but remains insufficient for pose estimation\cite{tian2022fcos,zhang2020atss}. Therefore, we propose a Texture-Shape-guided Feature Pyramid Networks (TS-FPN) to further integrate high and low-frequency features. Compared to FPN, in the top-down process, we do not simply use an upsampling and addition approach. As shown in Figure \ref{fig:TSFM}, our TS-FPN concatenates the upsampled low-frequency features $\mathcal{F} _{l}^{i}\in \mathbb{R} ^{C_i\times H_i\times W_i}$
 with the high-frequency features $\mathcal{F} _{h}^{i}\in \mathbb{R} ^{C_i\times H_i\times W_i}$
along the channel dimension. It then captures the spatial dependencies of \textit{H} and \textit{W} to generate attention weights. These attention weights are applied to $\mathcal{F} _{l}^{i}$ and $\mathcal{F} _{h}^{i}$ separately, and then the weighted high-frequency/low-frequency features are added back to $\mathcal{F} _{l}^{i}$/$\mathcal{F} _{h}^{i}$ to obtain enhanced features $\hat{\mathcal{F}}_{h}^{i}$/$\hat{\mathcal{F}}_{l}^{i}$. This enhancement process can be described as:
\begin{equation}
\begin{aligned}
    \hat{\mathcal{F}}_{h}^{i}=\mathcal{F} _{h}^{i}\oplus ( \mathcal{F} _{l}^{i}\otimes \sigma ( Conv( C\text{-}Pool( C( \mathcal{F} _{h}^{i},\mathcal{F} _{l}^{i} ) ) ) ) ) ,\\
    \hat{\mathcal{F}}_{l}^{i}=\mathcal{F} _{l}^{i}\oplus ( \mathcal{F} _{h}^{i}\otimes \sigma ( Conv( C\text{-}Pool( C( \mathcal{F} _{h}^{i},\mathcal{F} _{l}^{i} ) ) ) ) ) ,
    \end{aligned}
\end{equation}
where $\oplus$ refers to element-wise addition, $\otimes$ denotes element-wise multiplication, $\sigma\left( \cdot \right)$ represents the sigmoid function, $Conv\left( \cdot \right)$ stands for convolution with a kernel size of 7, $C\text{-}Pool\left( \cdot \right)$ indicates a channel pool, and $C\left( \cdot \right)$ means the concatenation along the channel dimension. 

After capturing the spatial dependencies, we rotate $\hat{\mathcal{F}}_{h}^{i}$/$\hat{\mathcal{F}}_{l}^{i}$ $90^{\circ}$ counterclockwise along the H-axis/W-axis\cite{misra2021triplet}. We then concatenate the features processed in the same way and generate attention weights between \textit{C} and \textit{W}, and between \textit{C} and \textit{H}, in parallel channel attention. These weights are used to reweight the corresponding enhanced features, followed by the inverse rotation. Finally, we take the average to obtain the final fused features $\mathcal{F} _{fu}^{i}$. The cross-channel feature fusion can be described as:
\begin{equation}
\begin{aligned}
\tilde{\mathcal{F}}_{1}^{i}=\hat{\mathcal{F}}_{h,r_{H}^{+}}^{i}\otimes \sigma ( Conv( Z\text{-}Pool( C( \hat{\mathcal{F}}_{h,r_{H}^{+}}^{i},\hat{\mathcal{F}}_{l,r_{H}^{+}}^{i} ) ) ) ) , \\
\tilde{\mathcal{F}}_{2}^{i}=\hat{\mathcal{F}}_{l,r_{H}^{+}}^{i}\otimes \sigma ( Conv( Z\text{-}Pool( C( \hat{\mathcal{F}}_{h,r_{H}^{+}}^{i},\hat{\mathcal{F}}_{l,r_{H}^{+}}^{i} ) ) )) ,\\
\tilde{\mathcal{F}}_{3}^{i}=\hat{\mathcal{F}}_{h,r_{W}^{+}}^{i}\otimes \sigma ( Conv( Z\text{-}Pool( C( \hat{\mathcal{F}}_{h,r_{W}^{+}}^{i},\hat{\mathcal{F}}_{l,r_{W}^{+}}^{i} ) ) ) ) , \\
\tilde{\mathcal{F}}_{4}^{i}=\hat{\mathcal{F}}_{l,r_{W}^{+}}^{i}\otimes \sigma ( Conv( Z\text{-}Pool( C( \hat{\mathcal{F}}_{h,r_{W}^{+}}^{i},\hat{\mathcal{F}}_{l,r_{W}^{+}}^{i} ) ) ) ) ,
\end{aligned}
\end{equation}
\vspace{-1.5em}
\begin{equation}
   \mathcal{F} _{fu}^{i}=\small{\frac{1}{4}}( \tilde{\mathcal{F}}_{1,r_{H}^{-}}^{i}+\tilde{\mathcal{F}}_{2,r_{H}^{-}}^{i}+\tilde{\mathcal{F}}_{3,r_{W}^{-}}^{i}+\tilde{\mathcal{F}}_{4,r_{W}^{-}}^{i} ) , 
\end{equation}
where $Z\text{-}Pool\left( \cdot \right)$ represents the max and average pooling along the 0th dimension\cite{misra2021triplet}, and the subscripts $r_{H}^{+}$, $r_{W}^{+}$, $r_{H}^{-}$, and $r_{W}^{-}$ indicate $90^{\circ}$ rotations: $r_{H}^{+}$ for counterclockwise along the \textit{H}, $r_{W}^{+}$ for counterclockwise along the \textit{W},$r_{H}^{-}$ for clockwise along the \textit{H},  and $r_{W}^{-}$ for counterclockwise along the \textit{W}.

Finally, our TS-FPN performs downsampling on the output side. It ultimately outputs feature maps at five scales, fused with shape and texture information, for further object detection and pose estimation.

\subsection{Pose Regression Strategy}
\label{sec:6d}
After obtaining the multi-scale features extracted by TS-FPN, we need to utilize these features for 6D pose estimation. For the rotation, our primary focus is on the object's appearance in the image, as rotation can significantly alter the object's appearance\cite{brownlee2019gentle}. Correspondingly, the estimation of the translation focuses more on the distance of the object's center point relative to the camera, as this directly affects the size of the object in the image\cite{amjoud2023object}. Therefore, the rotation and translation should be estimated using different strategies.

In the regression of rotation, all representations in four-dimensional or lower real Euclidean spaces are discontinuous. This means the widely used 3D and 4D representations (Euler angles and quaternions) are discontinuous, which poses challenges for neural network learning\cite{zhou2019continuity}. To address this, we adopted a higher-dimensional 6D representation. Compared to traditional representations, its main advantages are simplicity and continuity in most cases, and it has been proven effective in many studies\cite{wang2021gdr,zhou2019continuity}. In the 6D representation, a rotation is represented by two 3D vectors, which correspond to two orthogonal directions in 3D space. Our strategy is to directly regress the 6D vector $\boldsymbol{r}_{6d}=\left( r_1,r_2,r_3,r_4,r_5,r_6 \right)$ through the rotation head. Then, we can recover two 3D vectors from the 6D vector, i.e., $\boldsymbol{a}_1=\left( r_1,r_2,r_3 \right),$$
\boldsymbol{a}_2=\left( r_4,r_5,r_6 \right)$. Next, we construct three standard orthogonal bases $\boldsymbol{e}_1,$$\boldsymbol{e}_2,$ and $\boldsymbol{e}_3$ from these two 3D vectors. Finally, we use these three base vectors as the columns of the rotation matrix to obtain the final rotation matrix $\textbf{R}=\left[ \boldsymbol{e}_{1}^{T},\boldsymbol{e}_{2}^{T},\boldsymbol{e}_{3}^{T} \right]$. This process can be represented as:
\begin{equation}
    \boldsymbol{e}_1=\,\,\varPhi \left( \boldsymbol{a}_1 \right) ,
\\
\boldsymbol{e}_2=\varPhi \left( \boldsymbol{e}_1\times \boldsymbol{e}_2 \right) ,
\\
\boldsymbol{e}_3=\boldsymbol{e}_1\times \boldsymbol{e}_2,
\end{equation}
where $\varPhi\left( \cdot \right)$ denotes vector normalization, while the symbol $\times$ denotes the vector cross product.

After obtaining the rotation matrix, we use a geodesic loss shown in Equ. \ref{loss:rot} to train the rotation head\cite{mahendran20173d}. This formula quantifies the difference in angles between the predicted rotation and the actual rotation.
\begin{equation}
\mathcal{L} _{rot\_6d}=\mathrm{arc}\cos \left( \small{\frac{tr\left( \mathbf{R}_{gt}^{T}\mathbf{R}_{pred} \right) -1}{2}} \right) .
\label{loss:rot}
\end{equation}
To validate the effectiveness of this representation, we also experimented with the quaternion representation. Quaternions have two different representations for the same rotation. Therefore, we normalized both the predicted and actual quaternions by ensuring that the real part is non-negative. Additionally, we utilized a loss function shown in Equ. \ref{equ:quat}, which is essentially similar to the geodesic loss\cite{mahendran20173d}. The comparison of the effectiveness of the two representation methods is provided later in Sec. \ref{sec:experiment}.
\begin{equation}
\mathcal{L} _{rot\_quat}=2\mathrm{arc}\cos \left( \left| \left< \textbf{q}_{gt},\textbf{q}_{pred} \right> \right| \right) .
\label{equ:quat}
\end{equation}

For the translation, due to the translational invariance of convolutional operations, the network's prediction of absolute coordinates is not as effective as relative coordinates\cite{bukschat2020efficientpose}. Therefore, we did not directly regress the translation vector $
\mathbf{t}=\left( t_x, t_y, t_z \right) ^T
$ like in PoET. First, the transformation between the camera coordinate system and the pixel coordinate system can be expressed as $t_z\left( c_x,c_y,1 \right) ^T=K\left( t_x, t_y, t_z \right) ^T,$ where $c_x$ and $c_y$ represent the coordinates of the projected 3D point on the image, and \textit{K} represents the camera intrinsic parameters. Based on this, we decompose 
$\left( t_x, t_y, t_z \right) ^T$ into the projected points $\left( c_x, c_y \right) ^T$ and 
$t_z$. Then, we decompose $\left( c_x, c_y \right) ^T$ into the anchor point coordinates $
\left( a_x, a_y \right) ^T
$
and the relative coordinates $
\left( \Delta x, \Delta y \right) ^T
$. Therefore, we can obtain the translation by predicting the offset $
\left( \Delta x, \Delta y \right) ^T
$ and $t_z$, i.e., 
\begin{equation}
\begin{cases}
	t_x=\frac{\left( a_x+\Delta x-p_x \right) \cdot t_z}{f_x},\\
	t_y=\frac{\left( a_y+\Delta y-p_y \right) \cdot t_z}{f_y},\\
	t_z=t_z,\\
\end{cases}
\label{equ: trans}
\end{equation}
where $f_x$, $f_y$, $p_x$ and $p_y$ are obtained from the camera's intrinsic matrix $
K=\left( \begin{matrix}
	f_x&		0&		p_x\\
	0&		f_y&		p_y\\
	0&		0&		1\\
\end{matrix} \right) 
$. During training, we use the translation losses as
\begin{equation}
    \mathcal{L} _{tran_1}=\left\| \left( t_x,t_y \right) _{pred},\left( t_x,t_y \right) _{gt} \right\|_2,
\label{loss: tran1}
\end{equation}
\vspace{-1.5em}
\begin{equation}
    \mathcal{L} _{tran_2}=\left\| t_{z,pred},t_{z,gt} \right\| _1
.
\label{loss: tran2}
\end{equation}

\subsection{Pose Estimation Heads}
\label{sec:heads}

\begin{figure}
  \centering
  \begin{subfigure}[b]{0.225\textwidth}
    \centering
    \includegraphics[width=\linewidth]{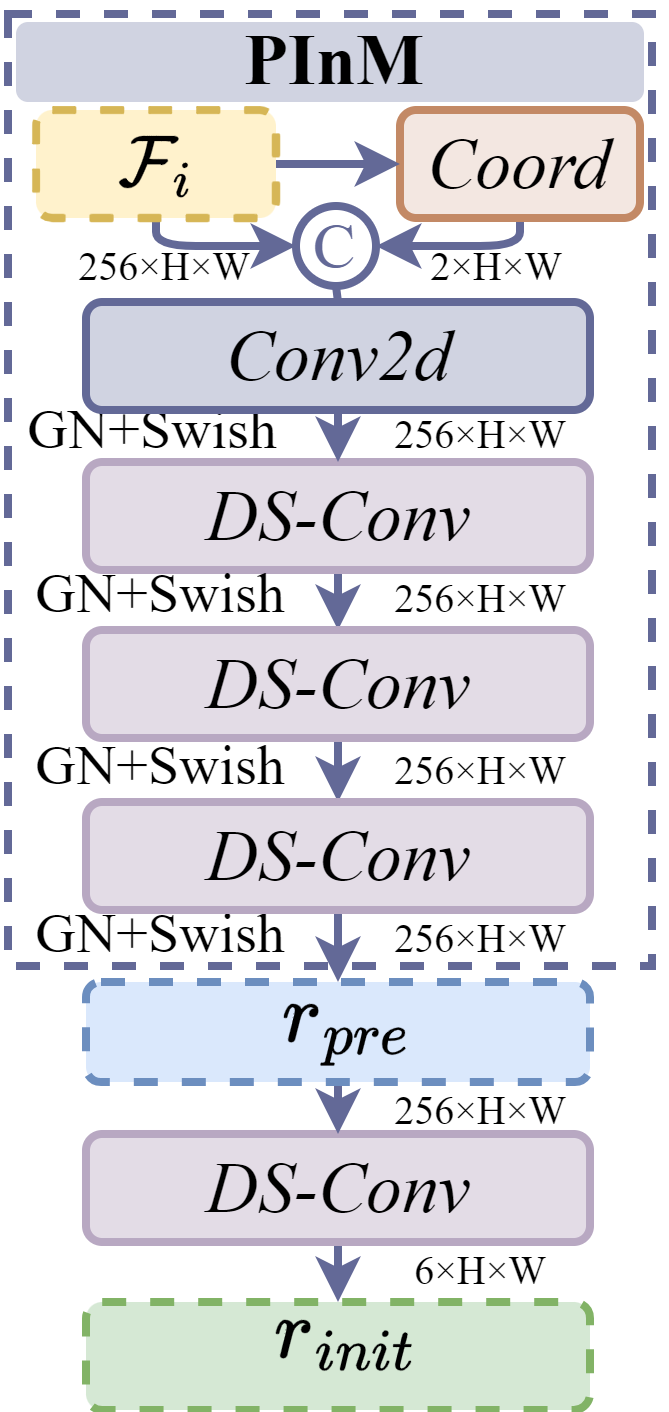}
    \caption{The part for initialization.}
    \label{fig:PInM}
  \end{subfigure}
  \hfill
  \begin{subfigure}[b]{0.245\textwidth}
    \centering
    \includegraphics[width=\linewidth]{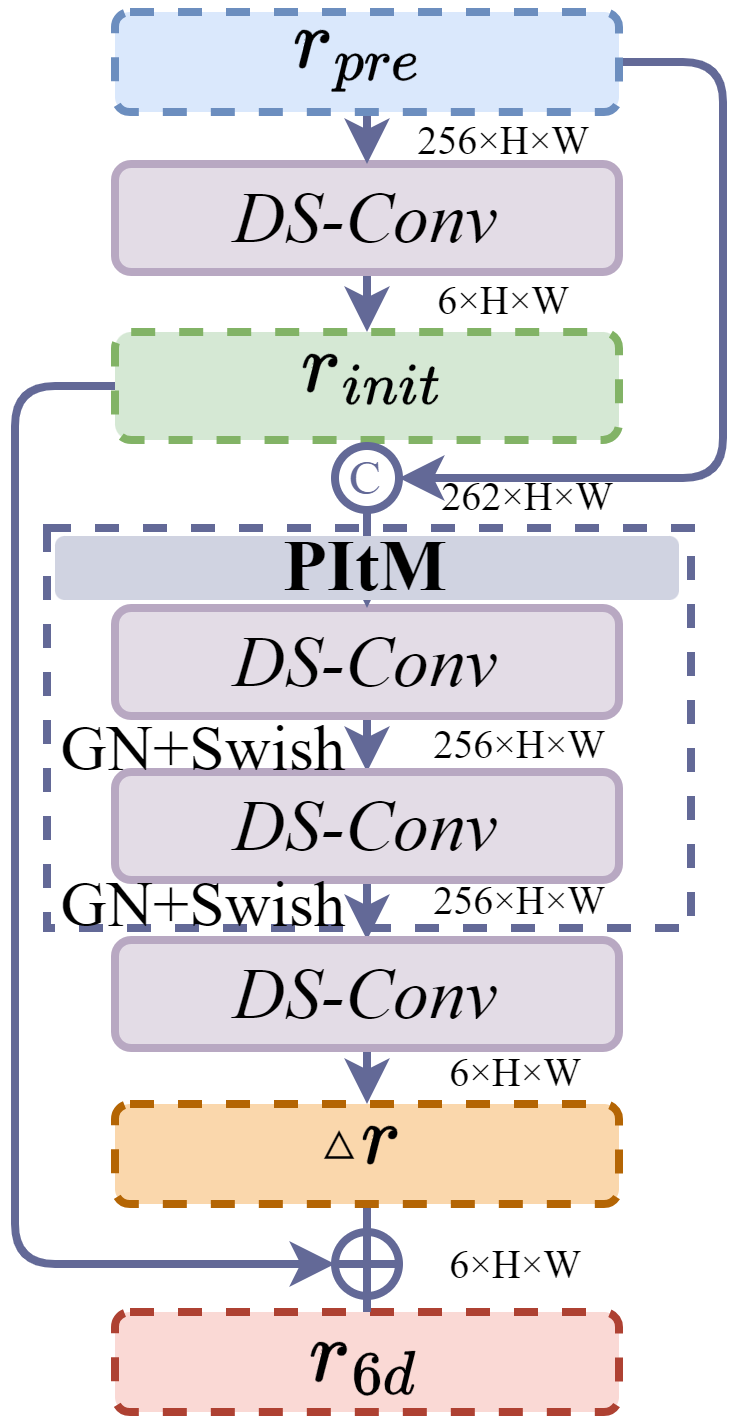}
    \caption{The part for iteration.}
    \label{fig:PItM}
  \end{subfigure}
  \caption{The structure of the rotation head.}
  \label{fig:rotation_head}
\end{figure}

In the previous section, we have established the regression strategies for rotation and translation. When implementing these strategies into the network structure, we need to regress $\textbf{r}_{6d}$ and $
\left( \Delta x,\Delta y,t_z \right) ^T
$ separately. The most straightforward approach is similar to the heads in the object detection task\cite{tian2022fcos,zhang2020atss}. Specifically, it involves the use of class head and bounding box head, as illustrated in Figure \ref{overview}. Four 3x3 convolutions are employed to further refine the feature maps produced by the TS-FPN. Lastly, a convolutional layer outputs the prediction results. However, compared to the object detection task, the 6D pose estimation task requires higher accuracy. Additionally, regressing the translation relies on coordinate features, which are difficult to extract with ordinary convolutions. Hence, we do not use a structure similar to the class head to regress rotation and translation. Inspired by EfficientPose\cite{bukschat2020efficientpose}, we adopt an iterative refinement approach to approximate the final result. Based on this, we design the rotation head and translation head. Taking the rotation head as an example, its structure is illustrated in Figure \ref{fig:rotation_head}.

The approximation approach is divided into two steps. Firstly, the pose initialization module (PInM) is utilized to refine the feature maps $
\mathcal{F} _i\in \mathbb{R} ^{C\times H\times W},i\in \left\{ 1,2,3,4,5 \right\} 
$ from TS-FPN. This refinement process results in obtaining $\boldsymbol{r}_{pre}$, which is then further processed to obtain $\boldsymbol{r}_{init}$. Specifically, the $x$ and $y$ coordinates of $\mathcal{F} _i$ are normalized to the range $\left[ 0,1 \right]$ and then mapped to $\left[ -1,1 \right]$. This process generates two channels with coordinate features, which are then concatenated with $\mathcal{F} _i$. The concatenated data passes through a standard convolutional layer. Subsequently, it is processed through three depthwise separable convolutional layers, along with group normalization and the Swish activation function, resulting in $\boldsymbol{r}_{pre}$. $\boldsymbol{r}_{pre}$, retaining the preliminary raw features, is utilized for further iterations. Finally, the number of channels is reduced to 6 using a depthwise separable convolutional layer to obtain $\boldsymbol{r}_{init}$. Thus, we have completed the addition of coordinate features and the initial utilization of overall features. Our first step can be represented as:
\begin{equation}
    \boldsymbol{r}_{pre}=DS\text{-}Conv^3\left( Conv\left( C\left( \mathcal{F} _i,Coord\left( \mathcal{F} _i \right) \right) \right) \right),
\end{equation}
\vspace{-1.5em}
\begin{equation}
  \boldsymbol{r}_{init}=DS\text{-}Conv\left( \boldsymbol{r}_{pre} \right),  
\end{equation}
where $DS\text{-}Conv\left( \cdot \right)$ refers to a depthwise separable convolution and $Coord\left( \cdot \right)$ indicates the operation of extracting coordinate information described above.

Secondly, we concatenate $\boldsymbol{r}_{pre}$ and $\boldsymbol{r}_{init}$ along the channel dimension. Then we feed the concatenated feature map into a pose iteration module (PItM) consisting of two depthwise separable convolutions. After that, we use another depthwise separable convolution to compress the channels, producing $\Delta \boldsymbol{r}$. At last, we add $\Delta \boldsymbol{r}$ to $\boldsymbol{r}_{init}$, obtaining the optimized $\boldsymbol{r}_{6d}$. The advantage of this structure is that it not only considers the initial rotation estimation but also integrates features from the early layers of the network. This helps to utilize more contextual information and offers a more comprehensive perspective for making predictions. Thus, we have completed one iteration. Our second step can be represented as:
\begin{equation}
    \Delta \boldsymbol{r}=DS\text{-}Conv^3\left( C\left( \boldsymbol{r}_{pre}, \boldsymbol{r}_{init} \right) \right) ,
\end{equation}
\vspace{-1.5em}
\begin{equation}
 \boldsymbol{r}_{6d}=\boldsymbol{r}_{init}+\Delta \boldsymbol{r}.   
\end{equation}

Our translation head and rotation head are essentially the same. The only difference is that we use two branches in the translation head, as shown in Figure \ref{overview}. This corresponds to $\left( \Delta x, \Delta y \right) ^T$ and $\textbf{t}_z$ in Sec. \ref{sec:6d}. At the end of the translation head, we concatenate the two branches along the channel dimension to output $
\left( \Delta x,\Delta y,t_z \right) ^T
$. Further, we can convert it back to the translation vector $\textbf{t}$.

Additionally, Our rotation and translation heads share weights across various scales, similar to FCOS's approach\cite{tian2022fcos}. The weight sharing leads to a considerable reduction in the number of parameters. This method ensures that as long as positive samples are present on one scale, weights on other scales can also be updated, despite having fewer positive samples. This helps mitigate the problem of uneven sample distribution across levels of heads.

\subsection{Positive Sample Selection}
\label{sec:sample}
\begin{figure}
  \centering
  \includegraphics[width=\linewidth]{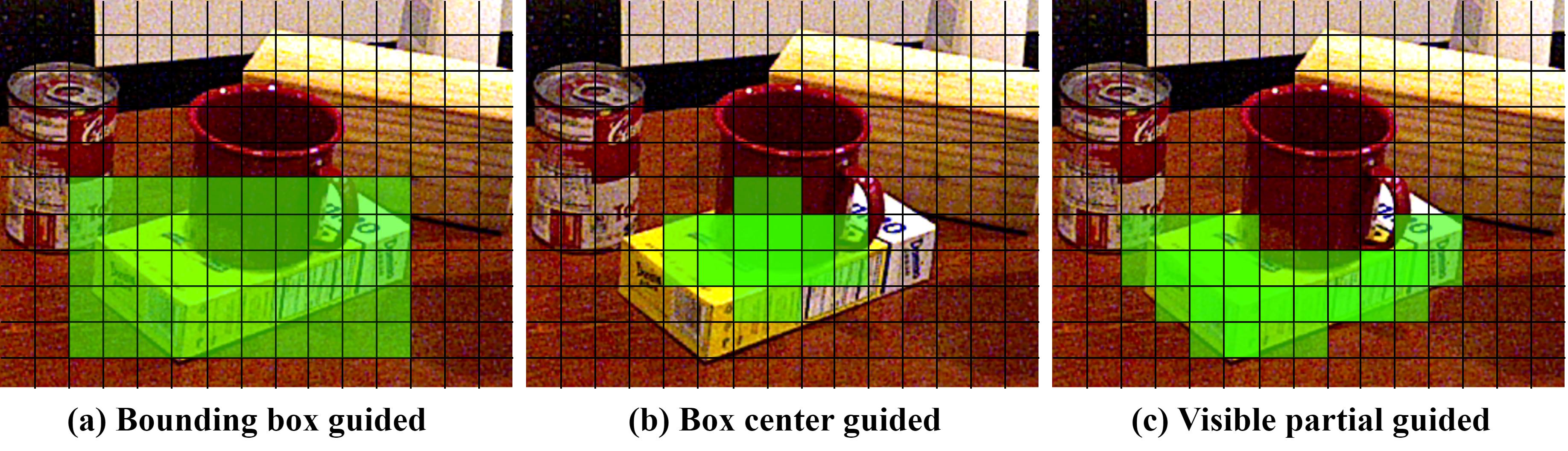}
  \caption{Schematic diagrams of the three different sampling strategies \cite{tian2022fcos,zhang2020atss,hai2023rigidity}. A darker green color indicates a higher probability of sampling.}
  \label{fig:3 sampling strategies}
\end{figure}

In this section, we will describe our approach to addressing occlusion in multi-object scenes. In single-stage object detectors, positive samples refer to sampling points linked to annotated instances of objects. These are utilized to depict the features of target categories during training. Typically, strategies for positive sample selection consider either all points within the GT bounding box or just those near the box center as positive samples\cite{tian2022fcos,zhang2020atss}. Hai et al. propose that it's more logical to use points from visible sections as positive samples for rigid objects\cite{hai2023rigidity}. In scenes with multiple objects, occlusion among objects is common, and the objects we study are all rigid. Therefore, during the training phase, we sample positive samples from the parts of the object that are visible. Different sampling strategies are illustrated in Figure \ref{fig:3 sampling strategies}.

\begin{figure}
  \centering
  \includegraphics[width=\linewidth]{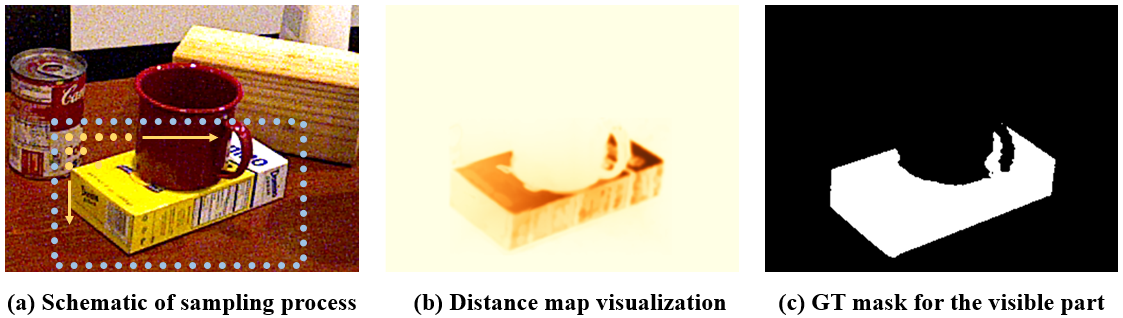}
  \caption{The schematic diagram of the sampling process from visible parts.}
  \label{fig: sampling progress}
\end{figure}

To implement the above sampling strategies, we first uniformly set boundary points $\mathbb{B} =\left\{ b_1, b_2, \cdots, b_m \right\} $ on the ground truth bounding box of the target object. Then, we calculate the foreground-background discrepancy $\mathcal{V} \left( p,b \right)$
 for each pixel point $p$ within the bounding box relative to the boundary point $b$. Finally, based on the foreground-background discrepancy, we determine the parts of the object that can be seen. The overall process is illustrated in the Figure \ref{fig: sampling progress}. We can see that the distance map and the ground truth visibility mask are essentially consistent. This means that during the process of positive sample sampling, we can achieve an effect similar to segmentation. 
 
 The principle of the above process is that boundary points are generally not part of the target object. The larger the discrepancy between a pixel point and a boundary point, the more likely the pixel point is part of the target object.  Specifically, we use the minimum barrier distance $\mathcal{D} \left( \mathcal{C}, l \right) $ related to the color difference \cite{hu2018minimum} and the Euclidean distance $d\left( p, d \right)$ to represent $\mathcal{V} \left( p,b \right)$. This can be formulated as, 
 \begin{equation}
     \mathcal{V} \left( p,b \right) =\min_{b\in \mathbb{B}} \min_{l\in \mathbb{L} _{\left\{ p,b \right\}}} \left[ \left( \frac{\mathcal{D} \left( \mathcal{C} , l \right)}{255} \right) ^2+\alpha \cdot d\left( p, d \right) \right] ,
 \end{equation}
 where $\mathbb{L} _{\left\{ p,b \right\}}$ represents the set of paths between $p$ and $b$, $l$  is one of these paths,  $\mathcal{C}$ represents the color brightness of $p$ and  $\alpha$ is a balance coefficient, set to 0.1 here, used to adjust the weight of color difference and distance. And $\mathcal{D} \left( \mathcal{C}, l \right)$ can be obtained by 
 \begin{equation}
     \mathcal{D} \left( \mathcal{C} ,l \right) =\,\,\underset{i=1}{\overset{3}{\max}}\left[ \max_{p_j\in l} \mathcal{C} _i\left( p_j \right) -\min_{p_j\in l} \mathcal{C} _i\left( p_j \right) \right].
 \end{equation}
In the formula, the outer $i$ represents one of the three channels of RGB, and the inner $p_j$ is a point on the path $l$. Furthermore, for the feature maps extracted by TS-FPN, we calculate the visibility probability for each cell by 
\begin{equation}
    \mathcal{P} \left( c \right) =\small{\frac{\overline{\mathcal{V} }\left( c \right)}{\max_{c\in \mathbb{C}} \overline{\mathcal{V} }\left( c \right)}},
\end{equation}
where $\overline{\mathcal{V} }\left( c \right)$ represents the average foreground-background discrepancy of pixels within a cell and $\mathbb{C}$ is the set of all cells on the feature maps. In our experiments, we set cells with  $\mathcal{P} \left( c \right) >0.25$ as positive samples. For more parameter settings, please refer to \cite{hai2023rigidity}.

\section{Experiments}
\label{sec:experiment}
In this section, we will first provide the implementation details of our experiment, descriptions of the datasets used, and the evaluation metrics. Then, we will compare our method with the state-of-the-art methods on two public datasets to demonstrate the superiority of SEMPose. Finally, we conduct ablation studies to prove the effectiveness of our specific designs.
\subsection{Experimental Setup}
\label{subsec: setup}
\textbf{Implementation Details.} Our algorithm is designed based on PyTorch. We choose ResNet-50 \cite{he2016deep} as the backbone network, utilize our TS-FPN for the neck, and design four heads corresponding to different tasks. Our TS-FPN is constructed from the last three layers of ResNet to extract features, generating multi-scale feature maps with 256 channels. In our network's training phase, we configure the batch size at 16 and opt for AdamW as the optimization algorithm, setting the initial learning rate to 4e-4. The optimizer's beta parameters are specified as $\left( 0.9,0.999 \right)$, coupled with a weight decay factor of 5e-4 and an epsilon of 1e-8. The OneCycle policy governs the learning rate adjustments\cite{smith2019super}, peaking at 4e-4. The training spans over 150 epochs, with the learning rate experiencing a linear ascent to its peak within the initial $0.5\%$ of the steps, followed by a linear decline throughout the subsequent steps. In the optimization configuration, we perform gradient clipping with a maximum norm of 35 using the L2 norm type. Most of our experiments are conducted on an RTX 4090D GPU and an AMD 3.1GHz CPU. Additionally, to ensure consistency with other methods, we measure the inference time on an RTX 3090 GPU and an Intel 2.8GHz CPU.
\begin{table*}[t]
\caption{\textbf{Comparison with State of the Art on LM-O.} We report the Recall of ADD(-S) in $\%$. P.E. indicates whether to use 1 model or N models to estimate N types of objects. (*) marks the symmetrical objects. (-) refers to unavailable results\cite{xiangposecnn,jantos2023poet,wang2021gdr,peng2019pvnet,zakharov2019dpod,li2018deepim,xu2024rnnpose}.}
\label{table: LM-O}
\begin{tabularx}{\textwidth}{X|*{6}{>{\centering\arraybackslash}X}*{3}{>{\centering\arraybackslash}X}}
\toprule
\multirow{-0.5}{*}{Method} & \multicolumn{6}{c|}{w/o Refinement} & \multicolumn{3}{c}{w/ Refinement} \\ \cline{2-10} 
                  & PoseCNN & PoET & GDR-Net & \multicolumn{1}{c|}{\textbf{ Ours}} & PVNet & \multicolumn{1}{c|}{GDR-Net} & DPOD & \multicolumn{1}{c|}{DeepIM} & RNNPose\\
                  \midrule
              P.E.    &  1 & 1  &  1 &  \multicolumn{1}{c|}{1} & N  & \multicolumn{1}{c|}{N}  & 1          &     \multicolumn{1}{c|}{1}  &N   \\ \midrule
                 Ape & 9.6  & 10.2  & \textbf{44.9}  & \multicolumn{1}{c|}{33.2}  &  15.8 &  \multicolumn{1}{c|}{46.8} &     -      & \multicolumn{1}{c|}{59.2}         &37.2\\
                  Can&  45.2 &  31.8 & 79.7  &  \multicolumn{1}{c|}{\textbf{85.2}} & 63.3  & \multicolumn{1}{c|}{90.8}  &  -         &     \multicolumn{1}{c|}{63.5} & 88.1    \\
                  Cat&  0.9 &  9.0 &  \textbf{30.6} &  \multicolumn{1}{c|}{30.1} & 16.7  & \multicolumn{1}{c|}{40.5}  &     -      &   \multicolumn{1}{c|}{26.2}     & 29.2  \\
                  Driller& 41.4  & 33.9  &  67.8 &\multicolumn{1}{c|}{\textbf{ 88.6}}  & 65.7  &  \multicolumn{1}{c|}{82.6} &    -       & \multicolumn{1}{c|}{55.6}      & 88.1   \\
                  Duck&  19.6 &  15.4 &  \textbf{40.0} & \multicolumn{1}{c|}{29.0}  & 25.2 & \multicolumn{1}{c|}{46.9}  &    -       & \multicolumn{1}{c|}{52.4}     &  49.2    \\
                   Eggbox*& 22.0  & 44.7  & 49.8  & \multicolumn{1}{c|}{\textbf{77.0} } &  50.2 &  \multicolumn{1}{c|}{54.2} &  -         &      \multicolumn{1}{c|}{63.0} &  67.0   \\
                    Glue*&  38.5 &   58.7&  73.7 &\multicolumn{1}{c|}{ 67.3}  & 49.6  & \multicolumn{1}{c|}{75.8}  &    -       &  \multicolumn{1}{c|}{71.7}     &63.8   \\
                  HoleP.&  22.1 &  24.7 & 62.7  & \multicolumn{1}{c|}{\textbf{71.4}}  &   36.1&  \multicolumn{1}{c|}{60.1} &  -         &       \multicolumn{1}{c|}{52.5} &   62.8  \\ \midrule
                  Mean&  24.9 &  28.5 & 56.1  & \multicolumn{1}{c|}{\textbf{60.2}}  & 40.8  &  \multicolumn{1}{c|}{62.2} &     47.3     &        \multicolumn{1}{c|}{55.5}  &   60.7\\ \bottomrule
\end{tabularx}
\end{table*}

\textbf{Datasets.} We use two core datasets from the BOP dataset, LM-O\cite{brachmann2014learning} and YCB-V\cite{xiangposecnn}, to evaluate our method. The LM-O dataset provides 6D pose annotations for 8 low-texture objects under various occlusion conditions, making it well-suited for testing an algorithm's occlusion handling capabilities. We train exclusively on Physically-Based Rendering (PBR) data and test on real-world data. YCB-V is a highly challenging dataset with varying lighting conditions, cluttered backgrounds, and image noise. It includes 21 commonly found household objects of different sizes, some of which exhibit high symmetry and varying degrees of occlusion among objects. The original YCB-V dataset comprises 92 video sequences, of which we use 80 sequences for real image training and the remaining 12 sequences for testing. Additionally, we also use the officially provided PBR data for training.

\textbf{Evaluation Metrics.} We use the commonly applied ADD(-S) metric for evaluation. ADD measures the average distance between corresponding model points transformed by ground truth and estimated poses, suitable for non-symmetric objects. ADD-S measures the average distance between each model point transformed by the estimated pose and the nearest model point transformed by the ground truth pose, suitable for symmetric objects. The specific formulas are as follows:
\begin{equation}
    ADD=\underset{\boldsymbol{x}\in \mathcal{M}}{avg}\left\| \left( \textbf{R}_{gt}\boldsymbol{x}+\boldsymbol{t}_{gt} \right) -\left( \textbf{R}_{pred}\boldsymbol{x}+\boldsymbol{t}_{pred} \right) \right\| ,
\label{metric: ADD}
\end{equation}
\vspace{-1.5em}
\begin{equation}
    ADD\text{-}S=\underset{\boldsymbol{x}_1\in \mathcal{M}}{avg}\underset{\boldsymbol{x}_2\in \mathcal{M}}{\min}\left\| \left( \textbf{R}_{gt}\boldsymbol{x}+\boldsymbol{t}_{gt} \right) -\left( \textbf{R}_{pred}\boldsymbol{x}+\boldsymbol{t}_{pred} \right) \right\| ,
\label{metric: ADD-S}
\end{equation}
where $\mathcal{M}$ refers to the set of 3D model points, $\textbf{R}_{gt}$ and $\boldsymbol{t}_{gt}$ represent the ground truth rotation and translation, while $\textbf{R}_{pred}$ and $\boldsymbol{t}_{pred}$ represent the predicted rotation and translation.
When evaluating on YCB-V, we also computed the AUC (area under curve) of ADD-S. Additionally, we calculated the geodesic distance as the rotation error and the L2 distance as the translation error for each object category, with the specific formulas as follows:
\begin{equation}
    error_{rot}=\mathrm{arc}\cos \left( \small{\frac{tr\left( \textbf{R}_{gt}^{T}\textbf{R}_{pred} \right) -1}{2}} \right) ,
\end{equation}
\vspace{-1.5em}
\begin{equation}
    error_{tran}=\left\| \boldsymbol{t}_{gt}-\boldsymbol{t}_{pred} \right\|.
\end{equation}

\subsection{Comparison with State of the Art}
\noindent\textbf{Results on LM-O.} Table. \ref{table: LM-O} shows our comparison results with state-of-the-art methods on LM-O\cite{xiangposecnn,jantos2023poet,wang2021gdr,peng2019pvnet,zakharov2019dpod,li2018deepim,xu2024rnnpose}. Among the single-model methods without using additional refinement, our SEMPose achieves the state-of-the-art results; compared to the methods that use 8 models for 8 objects, SEMPose also achieves similar accuracy. Moreover, SEMPose even outperforms some time-consuming methods that employ refinement.

\begin{table}[htbp]
\caption{\textbf{Comparison with State of the Art on YCB-V.} We report the results evaluated by AUC of ADD-S and AUC of ADD(-S). ADD-S means using symmetric metrics for all objects, while ADD(-S) is used only for symmetric objects.  P.E. indicates whether to use 1 model or N models to estimate N types of objects. (-) refers to unavailable results.}
\label{table: YCB-V}
\begin{tabularx}{\linewidth}{X|c|c|c|c}
\toprule
\multirow{2}{*}{Method} & \multirow{2}{*}{Refinement} & \multirow{2}{*}{P.E.} &  AUC of                    &              AUC of        \\
                  &                   &                   & ADD-S & ADD(-S) \\ 
\midrule
PoseCNN\cite{xiangposecnn} & & 1 & 75.9 & 61.3 \\
SilhoNet\cite{billings2019silhonet} & & 1 & 79.6 & - \\
PoET\cite{jantos2023poet} & & 1 & 87.1 & 70.1 \\
GDR-Net\cite{wang2021gdr} & & 1 & 89.1 & 80.2 \\
\textbf{Ours} & & 1 & \textbf{92.2} & \textbf{85.2} \\ 
\midrule
PVNet\cite{peng2019pvnet} & & N & - & 73.4 \\
GDR-Net\cite{wang2021gdr} & & N & 91.6 & 84.4 \\ 
\midrule
DeepIM\cite{li2018deepim} &\checkmark & 1 & 88.1 & 81.9 \\
CosyPose\cite{labbe2020cosypose} & \checkmark& 1 & 89.8 & 84.5 \\ 
\midrule
RNNPose\cite{xu2024rnnpose} &\checkmark & N & - & 83.1 \\ 
\bottomrule
\end{tabularx}
\end{table}

\noindent\textbf{Results on YCB-V.} In Table. \ref{table: YCB-V}, we present the comparison of SEMPose with other state-of-the-art methods on YCB-V. Among the single-model methods without using refinement, SEMPose continues to achieve the best results. For multi-model methods that use 21 models for prediction, our results remain competitive. Moreover, SEMPose also performs well in terms of accuracy compared to the time-consuming methods that employ refinement. 

Additionally, we report a comparison of the average rotation error and the average translation error for each category. The methods listed in the Table. \ref{table: error} are all single-model approaches for YCB-V. As the table shows, our method achieves lower translational prediction errors compared to other methods; the rotational prediction error is only bigger than that of SilhoNet, which employs rotation correction. Compared to GDR-Net, which performs second best in Table. \ref{table: YCB-V}, SEMPose reduces the translation prediction error by 2.47cm and the rotational error by 8.87 $^{\circ}$.
\begin{figure*}
  \centering
  \includegraphics[width=\linewidth]{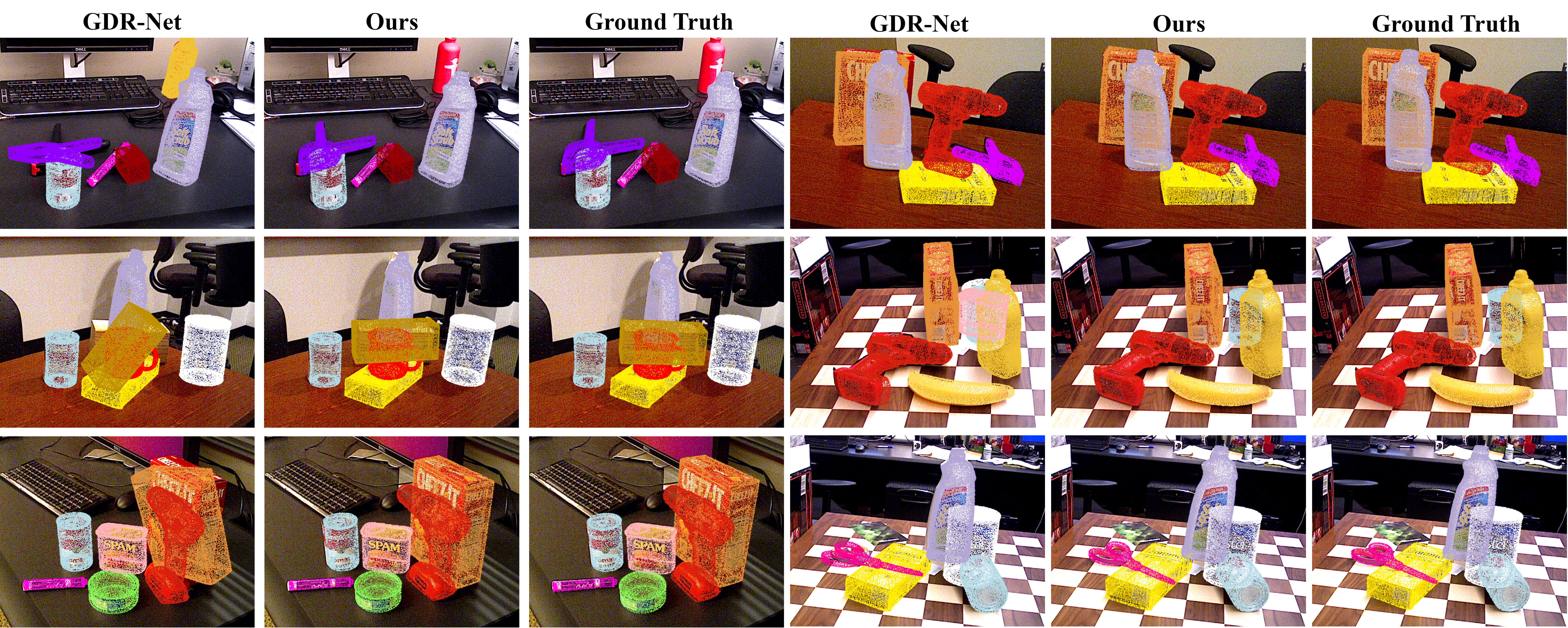}
  \caption{Qualitative comparison results on YCB-V.}
  \label{fig: results_compare}
\end{figure*}

\begin{table*}[]
\caption{\textbf{Comparison of average errors.} We report the average errors for translation and rotation. ($\dagger$) indicates the use of rotation correction from \textbf{SILHONET}\cite{}. (*) marks the symmetrical objects. The results of GDR-Net are obtained through the official single model\cite{wang2021gdr,xiangposecnn,jantos2023poet,billings2019silhonet}. }
\label{table: error}
\renewcommand{\arraystretch}{0.8}
\setlength{\tabcolsep}{2.8pt} 
\begin{tabular*}{\linewidth}{@{\extracolsep{\fill}}l|ccccc|ccccc}

\toprule
\multirow{2}{*}{Method} & \multicolumn{5}{c|}{Average Translation Error [cm]} & \multicolumn{5}{c}{Average Rotation Error [$^{\circ}$]} \\ 
\cmidrule(lr){2-6} \cmidrule(lr){7-11} 
                  & \makecell[cb]{PoseCNN} & \makecell[cb]{SilhoNet$^{\dagger}$ } & \makecell[cb]{PoET} & \makecell[cb]{GDR-Net} & \makecell[cb]{\textbf{Ours}} & \makecell[cb]{PoseCNN} & \makecell[cb]{SilhoNet$^{\dagger}$} & \makecell[cb]{PoET} & \makecell[cb]{GDR-Net} & \makecell[cb]{\textbf{Ours}} \\ \midrule
master chef can&3.29    &3.02    &2.26    &5.40    &\textbf{1.79}   &50.70    &\textbf{1.21}    &80.12    &48.16   &   22.33\\
cracker box&4.02    &5.24    &\textbf{3.14}    &5.09    &3.58   &\textbf{19.69}    &19.86    &21.87    &25.30   &  35.70 \\
sugar box&3.06    &2.10    &1.42    &5.23    &\textbf{1.15}   &9.29    &12.28    &\textbf{4.40}    &20.10   &   15.14\\
tomato soup can&3.02    &2.40    &1.62    &6.60    &\textbf{1.57}   &18.23    &\textbf{1.91}    &49.29    &30.23   &  17.00 \\
mustard bottle&1.72    &1.65    &1.42    &2.61    &\textbf{1.16}   &9.94    &\textbf{5.78}    &27.73    &9.16   &  13.78 \\
tuna fish can&2.41    &\textbf{1.57}    &1.79    &5.59    &1.76   &32.80    &\textbf{1.46}    &63.72    &30.53   &  40.70 \\
pudding box&3.69    &7.15    &1.94    &1.78    &\textbf{1.66}   &10.20    &20.95    &6.87    &\textbf{6.74}   &   10.21\\
gelatin box&2.49    &1.09    &1.41    &1.67    &\textbf{0.96}   &\textbf{5.25}    &12.52    &7.19    &7.88   &   35.35\\
potted meat can&3.65    &4.30    &\textbf{1.75}    &5.14    &1.93   &28.67    &7.27    &\textbf{6.75}    &18.24   &  14.67 \\
banana&2.43    &4.12    &1.95    &2.02    &\textbf{0.92}   &15.48    &16.29    &20.40    &9.16   &   \textbf{8.25}\\
pitcher base&4.43    &\textbf{1.31}    &1.55    &4.89    &1.53   &11.98    &\textbf{6.64}    &8.04    &30.17   &  10.16 \\
bleach cleanser&4.86    &3.60    &2.47    &7.52    &\textbf{3.19}   &20.85    &51.28    &21.93    &43.85   &  \textbf{17.49} \\
bowl*&5.23    &3.30    &1.76    &3.81    &\textbf{1.97}   &75.53    &49.95    &\textbf{25.71 }   &69.60   &  38.05 \\
mug&4.00    &2.61    &\textbf{1.85}    &2.77    &2.52   &19.44    &18.14    &\textbf{5.59 }   &7.75   &  15.88 \\
power drill&4.59    &6.77    &2.29    &4.80    &\textbf{1.85}   &9.91    &30.54    &\textbf{6.45}    &28.27   &   14.26\\
wood block*&6.34    &5.59    &4.75    &6.33    &\textbf{2.68}   &23.63    &25.52    &\textbf{14.32}    &79.75   &  42.37 \\
scissors&6.40    &9.91    &3.72    &3.69    &\textbf{2.06}   &43.98    &155.53    &\textbf{6.27}    &87.35    &   30.71\\
large marker&3.89    &3.24    &2.75    &4.45    & \textbf{1.76}  &92.44    &\textbf{10.44}    &25.91    &80.28   &  41.38 \\
large clamp*&9.79    &6.27    &\textbf{2.33}    &7.46    &4.50   &38.12    &\textbf{3.54}    &4.88    &12.01   &   54.25\\
extra large clamp*&8.36    &4.86    &\textbf{3.10 }   &5.83    &3.78   &34.18    &29.18    &\textbf{26.01}    &52.39   &   38.59\\
foam brick*&2.48    &3.98    &3.42    &2.36    &\textbf{1.75}   &22.67    &13.84    &36.34    &\textbf{4.41}   &   71.46\\ \midrule
Mean&4.16    &3.49   &2.12    &4.53    &\textbf{2.06}   &27.79    &\textbf{16.04}    &27.26    &33.39   &   24.52\\ \bottomrule
\end{tabular*}
\end{table*}

In Figure. \ref{fig: results_compare}, we present qualitative comparison results on YCB-V. We project the 3D model points onto the image using the predicted poses and differentiate objects with various colors. It is evident that our method effectively handles occlusion, a challenge that other single-model methods struggle with. For both GDR-Net and our SEMPose, a score threshold of 0.4 is used before plotting to remove low-confidence predictions. Despite this, GDR-Net frequently misidentifies background objects as targets or misclassifies targets as other objects. This is a common issue among single-model methods.

\begin{table*}[]
\caption{\textbf{Ablation Study on YCB-V.} Ablation on feature fusion, pose regression methods, pose head structures, and positive sample sampling strategies.}
\begin{tabular*}{\linewidth}{@{\extracolsep{\fill}}l|l|cccc}
\toprule
\multirow{2}{*}{Row} & \multirow{2}{*}{Method} & AUC of & AUC of & Avg. T. & Avg. R. \\
                  &                   & ADD-S & ADD(-S) & Error [cm]  & Error [$\circ$] \\ 
\midrule
A0&  \textbf{SEMPose (Ours)}               & \textbf{92.2} & \textbf{85.2} & \textbf{2.06} & \textbf{24.52} \\
B0& A0 $\rightarrow$ without fusion of texture and shape features                  &  85.7&73.2  &2.67  &52.48  \\
C0&A0: $\boldsymbol{r}_{6d}$ $\rightarrow$ quaternion                    &87.7  & 75.3 &2.69  & 50.45 \\
C1&A0: indirectly regress the translation vector $\rightarrow$ directly regression                 &  83.3&  67.7&3.51  &64.91  \\
C2&A0:  separate translation loss $\rightarrow$ combined translation loss                 & 91.7 & 82.7 & 2.15 & 27.58 \\
D0&A0: iterative refinement head $\rightarrow$  structure similar to the class head                   & 74.1 &56.7  &5.16  &72.37  \\ 
E0& A0: sampling from visible parts $\rightarrow$ from center parts                   & 84.2 &70.2  &3.12  &52.64  \\
\bottomrule
\end{tabular*}
\end{table*}

\subsection{Ablation Study on YCB-V}
We conduct ablation studies on the YCB-V dataset to demonstrate the effectiveness of our designs. All results are for the same set described in Sec. \ref{subsec: setup}.

\noindent\textbf{The fusion of texture and shape.} In Table. In row B0, we show the results without using the texture and shape fusion proposed in Sec. \ref{sec:tsfpn}. After removing the feature fusion, the prediction error for translation increased by 29.6$\%$, and the rotation error increased by 114.0$\%$. At the same time, the AUC of ADD-S and ADD(-S) decreased by 7.0$\%$ and 14.1$\%$, respectively. This indicates that by integrating texture and shape features, SEMPose can capture more accurate detail information, which is especially beneficial for rotation prediction.

\noindent\textbf{Pose regression strategies.} In rows C0 to C2, we show the outcomes of employing different pose regression strategies. Specifically: In row C0, we use quaternion instead of the 6D representation $\boldsymbol{r}_{6d}$ for representing rotation.  In row C1, we shift from regressing $\left( \Delta x, \Delta y \right) ^T$ and $t_z$ separately to directly regressing $\mathbf{t}=\left( t_x, t_y, t_z \right) ^T$. In row C2, we utilize the L2 distance between the predicted and actual translation vectors  for translation loss. Each adjustment leads to varying degrees of increased errors and reduced accuracy. It's critical to highlight that making adjustments for either rotation or translation alone influences the accuracy of the other value. This interdependence arises because the four heads share a common set of features derived from the backbone and neck of the network. Consequently, modifications targeted at one parameter can inadvertently alter the backbone and neck's parameters, thereby affecting the other value.

\noindent\textbf{Pose heads structure.} In row D0,  we substitute the iterative refinement heads described in Sec. \ref{sec:heads} with ones similar to the class head. This adjustment results in a significant loss in accuracy. The reason is that a simple head structure cannot effectively utilize the fused features from TS-FPN. Moreover, conventional convolutional layers do not proficiently extract coordinate features.

\noindent\textbf{Positive sample selection strategies.}  In row E0, we select positive samples for training from the center of the bounding box, leading to an expected decrease in accuracy. As mentioned in Sec. \ref{sec:sample}, occlusion is very common in pose estimation tasks. Significant occlusion results in the center of the bounding box no longer belonging to the target object. Sampling from visible parts can avoid this issue.

\subsection{Runtime Analysis}
\begin{figure}
  \centering
    \includegraphics[width=\linewidth]{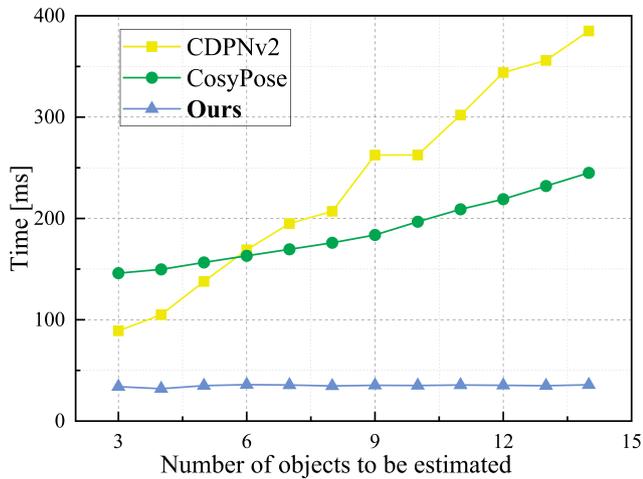}
    \caption{\textbf{Runtime comparison under the same conditions.} The running time of existing methods becomes longer as the number of objects to be estimated increases\cite{li2019cdpn, labbe2020cosypose}, whereas our method is unaffected. }
    \label{fig: run_time}
\end{figure}
On an RTX 3090 GPU and an Intel 2.8GHz CPU, inputting a $640\times480$ image, our SEMPose takes an average of 31ms to obtain the 6D poses of all target objects in the picture. This inference speed is highly competitive among mainstream methods. Moreover, as shown in Figure \ref{fig: run_time}, our inference time is independent of the number of objects in the image, which marks a significant difference from other methods.

\section{Conclusion and Future Work}
In this work, we propose the novel SEMPose, a single end-to-end network for multi-object pose estimation. In methods using only RGB images, SEMPose achieves state-of-the-art performance on the LM-O and YCB-V datasets. The key to success is that we employ a shape texture-guided feature processing strategy. This allows our SEMPose to better recognize individual objects from a multi-object scene. On the other hand, our iterative head structure also reduces the estimation error for rotations and translations. This guarantees the estimation accuracy of SEMPose. In addition, we sample positive samples from visible parts, which copes well with the occlusion problem in multi-object scenes. Based on the above, the SEMPose can predict the 6D poses of all target objects in a image accurately and in real time.

In the future, we plan to extend our work to category-level pose estimation, which involves predicting the poses of previously unseen objects within the same category. Additionally, we aim to explore the integration of depth information to further improve the robustness and accuracy of pose estimation, especially in challenging scenarios with severe occlusions and complex backgrounds.

\printcredits




\end{document}